\documentclass[10pt,twocolumn,letterpaper]{article}

\usepackage{cvpr}
\usepackage{times}
\usepackage{epsfig}
\usepackage{graphicx}
\usepackage{amsmath}
\usepackage{amssymb}
\usepackage{caption}
\usepackage{subcaption}
\usepackage{booktabs}
\usepackage{xcolor}
\usepackage{tabularx}
\usepackage{enumitem}

\usepackage[ruled,vlined,noend]{algorithm2e}
\SetAlCapNameFnt{\footnotesize}
\SetAlCapFnt{\footnotesize}

\usepackage[pagebackref=true,breaklinks=true,letterpaper=true,colorlinks,bookmarks=false]{hyperref}

\cvprfinalcopy 

\def\cvprPaperID{8111} 

\begin{document}

\title{What You See is What You Get: Exploiting Visibility for 3D Object Detection}

\author{
Peiyun Hu$^1$, Jason Ziglar$^2$, David Held$^1$, Deva Ramanan$^{1,2}$\\[.25em]
$^1$ Robotics Institute, Carnegie Mellon University\\
$^2$ Argo AI\\[.25em]
{\tt\small peiyunh@cs.cmu.edu, jziglar@argo.ai, dheld@andrew.cmu.edu, deva@cs.cmu.edu}\\
}

\maketitle


\begin{abstract}
Recent advances in 3D sensing have created unique challenges for computer vision. One fundamental challenge is finding a good representation for 3D sensor data. Most popular representations (such as PointNet) are proposed in the context of processing truly 3D data (e.g. points sampled from mesh models), ignoring the fact that 3D sensored data such as a LiDAR sweep is in fact 2.5D. We argue that representing 2.5D data as collections of $(x,y,z)$ points fundamentally destroys hidden information about freespace. In this paper, we demonstrate such knowledge can be efficiently recovered through 3D raycasting and readily incorporated into batch-based gradient learning. We describe a simple approach to augmenting voxel-based networks with visibility: we add a voxelized visibility map as an additional input stream. In addition, we show that visibility can be combined with two crucial modifications common to state-of-the-art 3D detectors: synthetic data augmentation of virtual objects and temporal aggregation of LiDAR sweeps over multiple time frames. On the NuScenes 3D detection benchmark, we show that, by adding an additional stream for visibility input, we can significantly improve the overall detection accuracy of a state-of-the-art 3D detector.
\end{abstract}

\section{Introduction}
What is a good representation for processing 3D sensor data? While this is a fundamental challenge in machine vision dating back to stereoscopic processing, it has recently been explored in the context of deep neural processing of 3D sensors such as LiDARs. Various representations have been proposed, including graphical meshes~\cite{bronstein2017geometric}, point clouds~\cite{qi2017pointnet}, voxel grids~\cite{zhou2018voxelnet}, and range images~\cite{meyer2019lasernet}, to name a few.

{\bf Visibility:} We revisit this question by pointing out that 3D sensored data, is infact, not fully 3D! Instantaneous depth measurements captured from a stereo pair, structured light sensor, or LiDAR undeniably suffer from occlusions: once a particular scene element is measured at a particular depth, visibility ensures that all other scene elements behind it along its line-of-sight are occluded. Indeed, this loss of information is one of the fundamental reasons why 3D sensor readings can often be represented with 2D data structures - e.g., 2D range image. From this perspective, such 3D sensored data might be better characterized as ``2.5D"~\cite{marr1978representation}.

{\bf 3D Representations:} We argue that representations for processing LiDAR data should embrace visibility, particularly for applications that require instantaneous understanding of freespace (such as autonomous navigation). However, most popular representations are based on 3D point clouds (such as PointNet~\cite{qi2017pointnet,lang2019pointpillars}). Because these were often proposed in the context of truly 3D processing (e.g., of 3D mesh models), they do not exploit visibility constraints implicit in the sensored data (Fig.~\ref{fig:splash}). Indeed, representing a LiDAR sweep as a collection of $(x,y,z)$ points fundamentally {\em destroys} such visibility information if normalized (e.g., when centering point clouds).

{\bf Occupancy:} By no means are we the first to point out the importance of visibility. In the context of LiDAR processing, visibility is well studied for the tasks of map-building and occupancy reasoning~\cite{thrun1996integrating,hornung2013octomap}. However, it is not well-explored for object detection, with a couple of notable exceptions. \cite{yapo2008probabilistic} builds a probabilistic occupancy grid and performs template matching to directly estimate the probability of an object appearing at each discretized location. However, this approach requires knowing surface shape of object instances before hand, therefore it is not scalable. \cite{chen20173d} incorporates visibility as part of an energy-minimization framework for generating 3D object proposals. In this work, we demonstrate that deep architectures can be simply augmented to exploit visibility and freespace cues.

{\bf Range images:} Given our arguments above, one solution might be defining a deep network on 2D range image input, which {\em implicitly} encodes such visibility information. Indeed, this representation is popular for structured light ``RGBD" processing~\cite{kim20113d,eitel2015multimodal}, and has also been proposed for LiDAR~\cite{meyer2019lasernet}. However, such representations do not seem to produce state-of-the-art accuracy for 3D object understanding, compared to 3D voxel-based or top-down, bird's-eye-view (BEV) projected grids. We posit that convolutional layers that operate along a depth dimension can reason about uncertainty in depth. To maintain this property, we introduce simple but novel approaches that directly augment state-of-the-art 3D voxel representations with visibility cues.

{\bf Our approach:} We propose a deep learning approach that efficiently augments point clouds with visibility. Our specific constributions are three-fold; (1) We first (re)introduce raycasting algorithms that effciently compute on-the-fly visibility for a voxel grid. We demonstrate that these can be incorporated into batch-based gradient learning. (2) Next, we describe a simple approach to augmenting voxel-based networks with visibility: we add a voxelized visibility map as an additional input stream, exploring alternatives for early and late fusion; (3) Finally, we show that visibility can be combined with two crucial modifications common to state-of-the-art networks: synthetic data augmentation of virtual objects, and temporal aggregation of LiDAR sweeps over multiple time frames. We show that visibility cues can be used to better place virtual objects. We also demonstrate that visibility reasoning over multiple time frames is akin to online occupancy mapping.

\section{Related Work}

\subsection{3D Representations}

{\bf Point representation:} Most classic works on point representation employ \textit{hand-crafted} descriptors and require robust estimates of local surface normals, such as spin-images~\cite{johnson1999using} and Viewpoint Feature Histograms (VFH)~\cite{rusu2010fast}. Since PointNet~\cite{qi2017pointnet}, there has been a line of work focuses on learning better point representation, including PointNet++\cite{qi2017pointnet++}, Kd-networks~\cite{klokov2017escape}, PointCNN~\cite{li2018pointcnn}, EdgeConv~\cite{wang2019dynamic}, and PointConv~\cite{wu2019pointconv} to name a few. Recent works on point-wise representation tend not to distinguish between \textit{reconstructed} and \textit{measured} point clouds. We argue that when the input is a \textit{measured} point cloud, e.g. a LiDAR sweep, we need to look beyond points and reason about visibility that is hidden within points.

{\bf Visibility representation:} Most research on visibility representation has been done in the context of robotic mapping. For example, Buhmann et al.~\cite{buhmann1995mobile} estimates a 2D probabilistic occupancy map from sonar readings to navigate the mobile robot and more recently Hornung et al.~\cite{hornung2013octomap} have developed Octomap for general purpose 3D occupancy mapping. Visibility through raycasting is at the heart of developing such occupancy maps. Despite the popularity, such visibility reasoning has not been widely studied in the context of object detection, except a notable exception of \cite{yapo2008probabilistic}, which develops a probabilistic framework based on occupancy maps to detect objects with known surface models.

\subsection{LiDAR-based 3D Object Detection}

\begin{figure*}
  \centering
  \includegraphics[width=\linewidth]{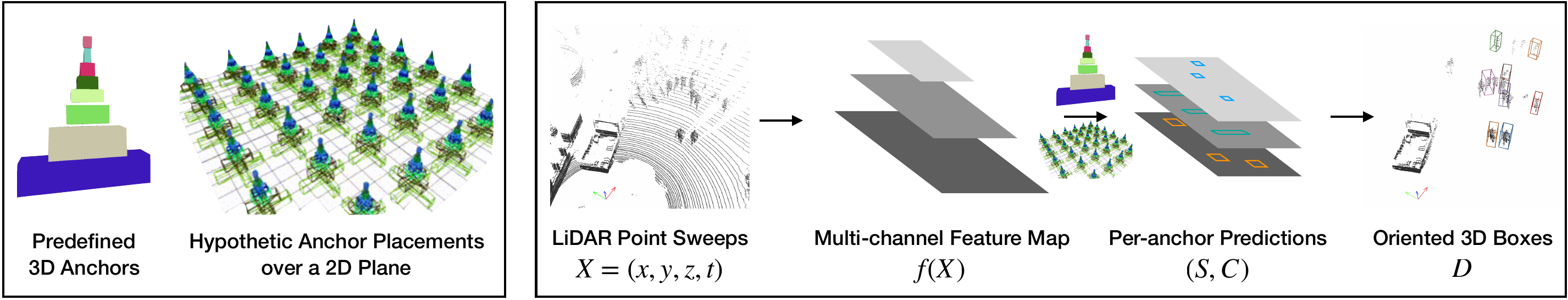}
  \caption{Overview of a general 3D detection framework, designed to solve 3D detection as a bird's-eye-view (BEV) 2D detection problem. The framework consists of two parts: anchors (\textbf{left}) and network (\textbf{right}). We first define a set of 3D anchor boxes that match the average box shape of different object classes. Then we hypothesize placing each anchor at different spatial locations over a ground plane. We learn a convolutional network to predict confidence and adjustments for each anchor placement. Such predictions are made based on 2D multi-channel feature maps, extracted from the input 3D point cloud. The predictions for each anchor consist of a confidence score $S$ and a set of coefficients $C$ for adjusting the anchor box. Eventually, the framework produces a set of 3D detections with oriented 3D boxes. \vspace{-1em}} 
  \label{fig:framework}
\end{figure*}

{\bf Initial representation:} We have seen LiDAR-based object detectors built upon range images, bird's-eye-view feature maps, raw point clouds, and also voxelized point clouds. One example of a range image based detector is LaserNet~\cite{meyer2019lasernet}, which treats each LiDAR sweep as a cylindrical range image. Examples of bird-eye-view detectors include AVOD~\cite{ku2018joint}, HDNet~\cite{yang2018hdnet}, and Complex-YOLO~\cite{simon2018complex}. One example that builds upon raw point clouds is PointRCNN~\cite{shi2019pointrcnn}. Examples of voxelized point clouds include the initial VoxelNet\cite{zhou2018voxelnet}, SECOND~\cite{yan2018second}, and PointPillars~\cite{lang2019pointpillars}. Other than~\cite{yapo2008probabilistic}, we have not seen a detector that uses visibility as the initial representation.

{\bf Object augmentation:} Yan et al.~\cite{yan2018second} propose a novel form of data augmentation, which we call \textit{object augmentation}. It copy-pastes object point clouds from one scene into another, resulting in new training data. This augmentation technique improves both convergence speed and final performance and is adopted in all recent state-of-the-art 3D detectors, such as PointRCNN~\cite{shi2019pointrcnn} and PointPillars~\cite{lang2019pointpillars}. For objects captured under the same sensor setup, simple copy-paste preserves the relative pose between the sensor and the object, resulting in approximately correct return patterns. However, such practice often inserts objects regardless of whether it violates the scene visibility. In this paper, we propose to use visibility reasoning to maintain correct visibility while augmenting objects across scenes.

{\bf Temporal aggregation:} When learning 3D object detectors over a series of LiDAR sweeps, it is proven helpful to aggregate information across time. Luo et al.~\cite{luo2018fast} develop a recurrent architecture for detecting, tracking, and fore-casting objects on LiDAR sweeps. Choy et al.~\cite{choy20194d} propose to learn spatial-temporal reasoning through 4D ConvNets. Another technique for temporal aggregation, first found in SECOND~\cite{yan2018second}, is to simply aggregate point clouds from different sweeps while preserving their timestamps relative to the current one. These timestamps are treated as additional per-point input feature along with $(x,y,z)$ and fed into point-wise encoders such as PointNet. We explore temporal aggregation over visibility representations and point out that one can borrow ideas from classic robotic mapping to integrate visibility representation with learning.

\section{Exploit Visibility for 3D Object Detection}

Before we discuss how to integrate visibility reasoning into 3D detection, we first introduce a general 3D detection framework. Many 3D detectors have adopted this framework, including AVOD~\cite{ku2018joint}, HDNet~\cite{yang2018hdnet}, Complex-YOLO~\cite{simon2018complex}, VoxelNet~\cite{zhou2018voxelnet}, SECOND~\cite{yan2018second}, and PointPillars~\cite{lang2019pointpillars}. Among the more recent ones, there are two crucial innovations: (1) object augmentation by inserting rarely seen (virtual) objects into training data and (2) temporal aggregation of LiDAR sweeps over multiple time frames.

We integrate visibility into the aforementioned 3D detection framework. First, we (re)introduce a raycasting algorithm that efficiently computes visibility. Then, we introduce a simple approach to integrate visibility into the existing framework. Finally, we discuss visibility reasoning within the context of object augmentation and temporal aggregation. For object augmentation, we modify the raycasting algorithm to make sure visibility remains valid while inserting virtual objects. For temporal aggregation, we point out that visibility reasoning over multiple frames is akin to online occupancy mapping.

\subsection{A General Framework for 3D Detection}
\label{sec:general}

{\bf Overview:} We illustrate the general 3D detection framework in Fig.~\ref{fig:framework}. Please refer to the caption. We highlight the fact that once the input 3D point cloud is converted to a multi-channel BEV 2D representation, we can make use of standard 2D convolutional architectures. We later show that visibility can be naturally incorporated into this 3D detection framework.

{\bf Object augmentation:} Data augmentation is a crucial ingredient of contemporary training protocols. Most augmentation strategies perturb coordinates through random transformations (e.g. translation, rotation, flipping)~\cite{ku2018joint,qi2018frustum}. We focus on \textit{object augmentation} proposed by Yan et al.~\cite{yan2018second}, which copy-pastes (virtual) objects of rarely-seen classes (such as buses) into LiDAR scenes. Our ablation studies (g$\rightarrow$i in Tab.~\ref{tab:ablation}) suggest that it dramatically improves vanilla PointPillars by an average of \textbf{+9.1\%} on the augmented classes.

{\bf Temporal aggregation:} In LiDAR-based 3D detection, researchers have explored various strategies for temporal reasoning. We adopt a simple method that aggregates (motion-compensated) points from different LiDAR sweeps into a single scene~\cite{yan2018second,caesar2019nuscenes}. Importantly, points are augmented with an additional channel that encodes the relative timestamp $(x,y,z,t)$. Our ablation studies (g$\rightarrow$j in Tab.~\ref{tab:ablation}) suggest that temporal aggregation dramatically improves the overall mAP of vanilla PointPillars model by \textbf{+8.6\%}.

\subsection{Compute Visibility through Raycasting}
{\bf Physical raycasting in LiDAR:} Each LiDAR point is generated through a physical raycasting process. To generate a point, the sensor emits a laser pulse in a certain direction. The pulse travels through air forward and back after hitting an obstacle. Upon its return, one can compute a 3D coordinate derived from the direction and the time-of-flight. However, coordinates are by no means the only information offered by such active sensing. Crucially, it also provides estimates of freespace along the ray of the pulse. 

{\bf Simulated LiDAR raycasting:} By exploiting the causal relationship between freespace and point returns - points lie along the ray where freespace ends, we can re-create the instantaneous visibility encountered at the time of LiDAR capture. We do so by drawing a line segment from the sensor origin to a 3D point. We would like to use this line segment to define freespace across a discretized volume, e.g. a 3D voxel grid. Specifically, we compute all voxels that intersect this line segment. Those that are encountered along the ray are marked as free, except the last voxel enclosing the 3D point is marked as occupied. 
This results in a visibility volume where all voxels are marked as occupied, free, or unknown (default). We will integrate the visibility volume into the general detection framework (Fig.~\ref{fig:framework}) in the form of a \textit{multi-channel 2D feature map} (e.g. a RGB image is an example with 3 channels) where visibility along the vertical dimension (z-axis) is treated as different channels. 

{\bf Efficient voxel traversal:} Visibility computation must be extremely efficient. 
Many detection networks exploit sparsity in LiDAR point clouds: PointPillars\cite{lang2019pointpillars} process only non-empty pillars (about 3\%) and SECOND~\cite{yan2018second} employs spatially sparse 3D ConvNets. Inspired by these approaches, we exploit sparsity through an efficient voxel traversal algorithm~\cite{amanatides1987fast}. For any given ray, we need traverse only a sparse set of voxels along the ray. Intuitively, during the traversal, the algorithm enumerates over the six axis-aligned faces of the current voxel to determine which is intersected by the exiting ray (which is quite efficient). It then simply advances to the neighboring voxel with a shared face. The algorithm begins at the voxel at the origin and terminates when it encounters the (precomputed) voxel occupied by the 3D point. This algorithm is linear in the grid dimension, making it quite efficient. Given an \textit{instantaneous} point cloud, where points are captured at the same timestamp, we perform raycasting 
from the origin to each point and aggregate voxels' visibility afterwards. 
To reduce discretization effects during aggregation, we follow best-practices outlined in Octomap (Sec. 5.1 in~\cite{hornung2013octomap}). 

\begin{figure*}[t]
  \centering
  \includegraphics[trim=0 350 0 100, clip, width=.49\linewidth]{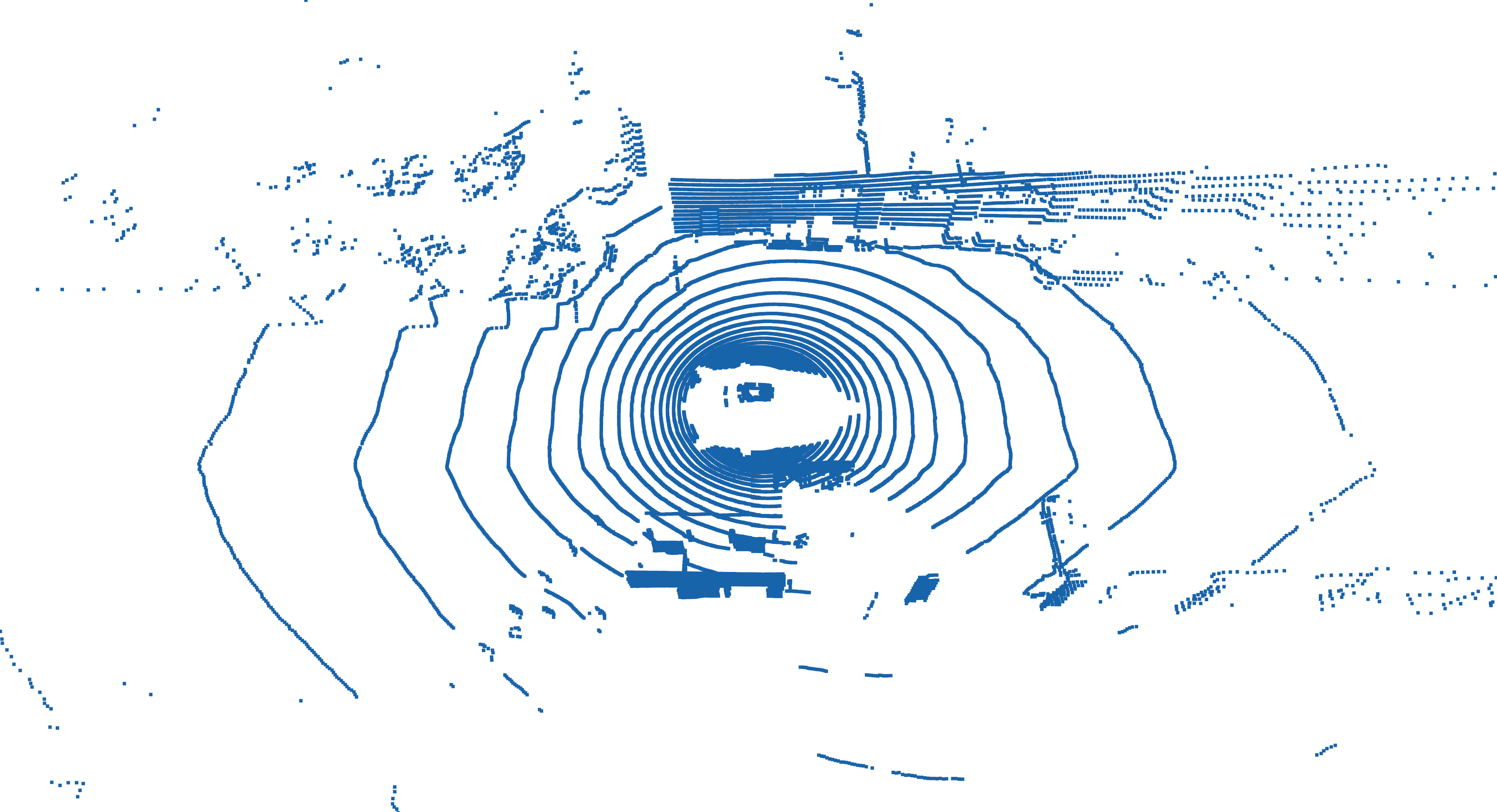}
  \includegraphics[trim=0 350 0 100, clip, width=.49\linewidth]{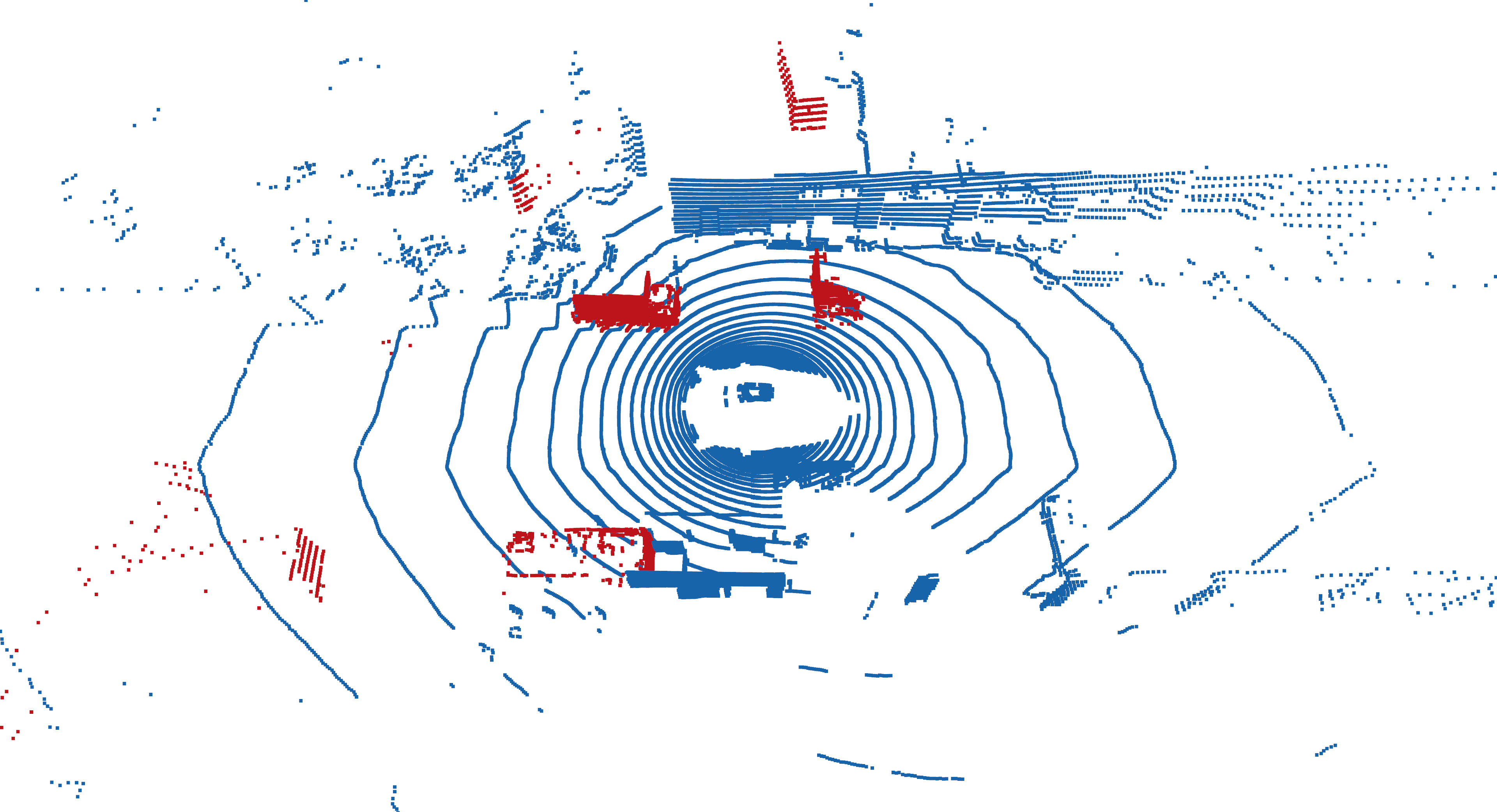}\\
  (a) original \hspace{20em} (b) naive \\[.5em]
  \includegraphics[trim=0 350 0 100, clip, width=.49\linewidth]{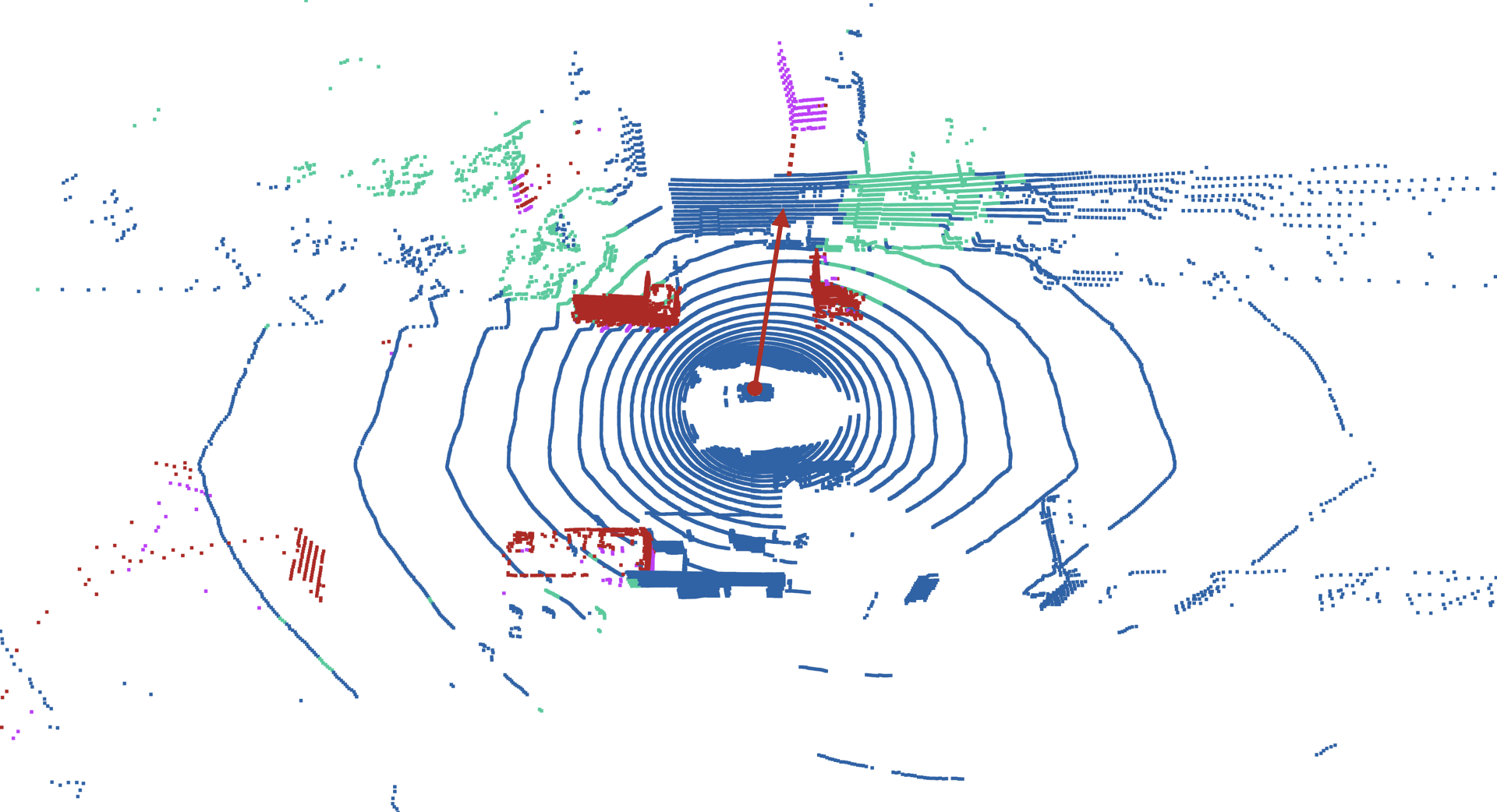}
  \includegraphics[trim=0 350 0 100, clip, width=.49\linewidth]{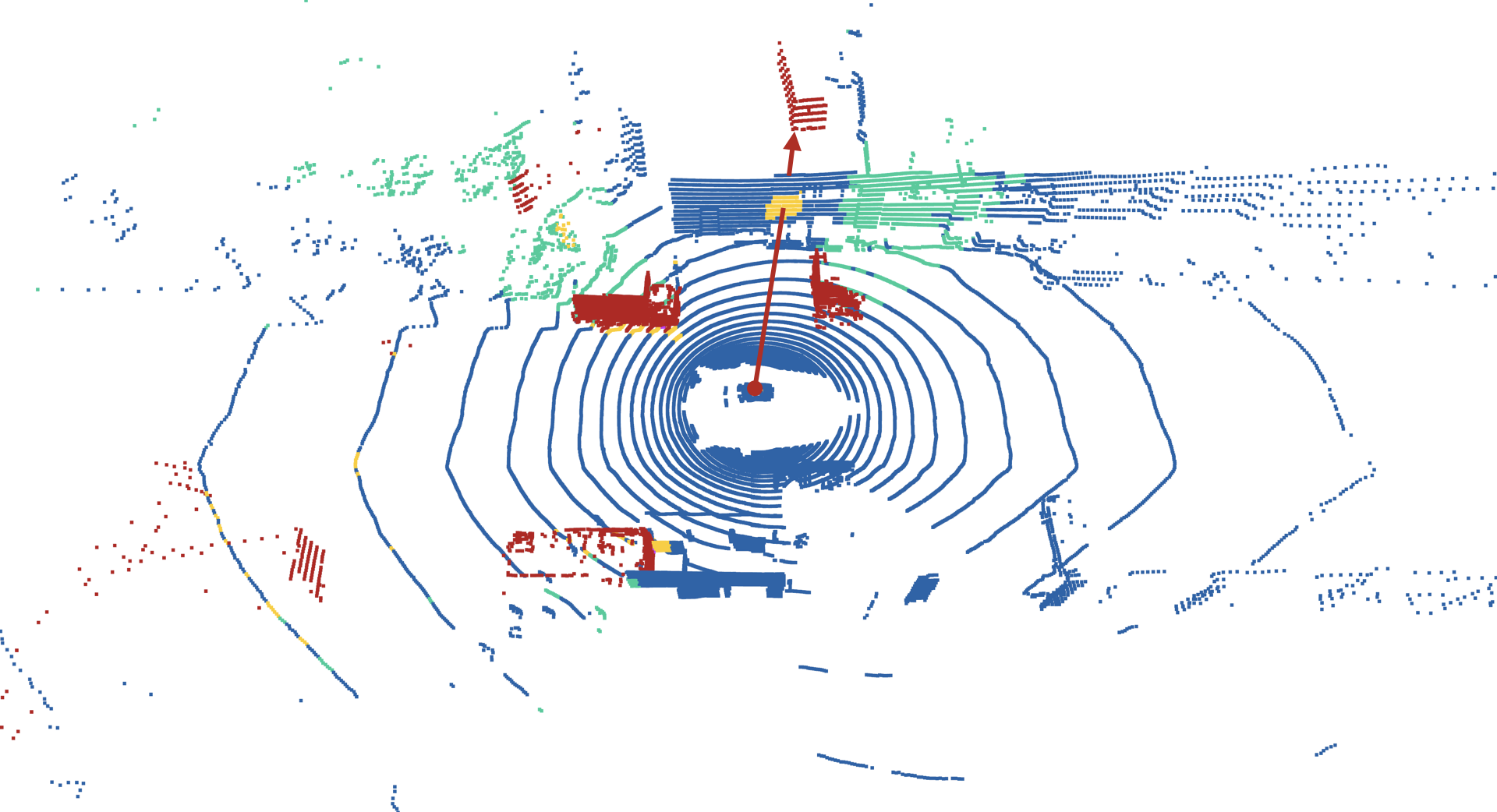}\\
  (c) culling \hspace{20em} (d) drilling \\
  \caption{
    Different types of object augmentation we can do through visibility reasoning.
    In (a), we show the original LiDAR point cloud.
    In (b), we naively insert new objects (red) into the scene.
    Clearly, the naive strategy may result in inconsistent visibility.
    Here, a trailer is inserted behind a wall that should occlude it.
    We use raycasting as a tool to ``rectify'' the LiDAR sweep.
    In (c), we illustrate the culling strategy, where we remove virtual objects that are occluded (purple). In practice, this may excessively remove augmented objects.
    In (d), we visualize the drilling strategy, where we remove points from the original scene that occlude the virtual objects. Here, a small piece of wall is removed (yellow).
  }
  \label{fig:augmentation}
\end{figure*}

{\bf Raycasting with augmented objects:} Prior work augments virtual objects while ignoring visibility constraints, producing LiDAR sweeps with inconsistent visibility (e.g., by inserting an object behind a wall that should occlude it - Fig.~\ref{fig:augmentation}-(b)). We can use raycasting as a tool to ``rectify" the LiDAR sweep. Specifically, we might wish to remove virtual objects that are occluded (a strategy we term \textit{culling} - Fig.~\ref{fig:augmentation}-(c)). Because this might excessively decrease the number of augmented objects, another option is to remove points from the original scene that occlude the inserted objects (a strategy we term \textit{drilling} - Fig.~\ref{fig:augmentation}-(d)).

Fortunately, as we show in Alg.~\ref{alg:raycast}, both strategies are efficient to implement with simple modifications to the vanilla voxel traversal algorithm. 
We only have to change the terminating condition of raycasting from arriving at the end point of the ray to hitting a voxel that is 
{\footnotesize $\mathtt{BLOCKED}$}. For \textit{culling},
when casting rays from the original scene, we set voxels occupied by virtual objects as 
{\footnotesize $\mathtt{BLOCKED}$}; when casting rays from the virtual objects, we set voxels occupied in original scenes as 
{\footnotesize $\mathtt{BLOCKED}$}.
As a result, points that should be occluded will be removed. For \textit{drilling}, we allow rays from virtual objects to pass through voxels occupied in the original scene.  

\begin{algorithm}[t]
  \scriptsize
  \SetAlgoLined
  \KwIn{mode $\mathbf{m}$, sensor origin $\mathbf{s}$, original points $\mathbf{P}$, augmented points $\mathbf{Q}$}
  \KwOut{occupancy grid $\mathbf{O}$}
  \textbf{Initialize:} $\mathbf{O}[:] \gets \mathtt{UNKNOWN}$ \;
  \tcc{Raycast P with Q as a ray stopper}
  Compute $\mathbf{B}$ such that $\forall \mathbf{q} \textrm{ in } \mathbf{Q}, \mathbf{B}[v_\mathbf{q}] \gets \mathtt{BLOCKED}$ \;
  \For{$\mathbf{p} \textrm{ in } \mathbf{P}$}{
    $v \gets v_{\mathbf{s}}$\tcc*[r]{$v_s$: sensor voxel}
    \While{$v \neq v_{\mathbf{p}}$}{
      $v \gets \mathtt{next\_voxel}(v, \mathbf{p} - \mathbf{s}$) \;
      \If{$\mathbf{B}[v] = \mathtt{BLOCKED}$}{
        break\tcc*[r]{stop the ray}
      }
      \eIf{$v = v_{\mathbf{p}}$}{
        $\mathbf{O}[v] \gets \mathtt{OCCUPIED}$ \;
      }{
        $\mathbf{O}[v] \gets \mathtt{FREE}$ \;
      }
    }
  }
  \tcc{Raycast Q with P as a ray stopper}
  Compute $\mathbf{B}$ such that $\forall \mathbf{q} \textrm{ in } \mathbf{Q}, \mathbf{B}[v_\mathbf{q}] \gets \mathtt{BLOCKED}$ \;
  \For{$\mathbf{q} \textrm{ in } \mathbf{Q}$}{
    $v \gets v_{\mathbf{s}}$\tcc*[r]{$v_s$: sensor voxel}
    \While{$v \neq v_{\mathbf{q}}$}{
      $v \gets \mathtt{next\_voxel}(v, \mathbf{q} - \mathbf{s}$) \;
      \If{$\mathbf{B}[v] = \mathtt{BLOCKED}$}{
        \uIf{$\mathbf{m}=\mathtt{CULLING}$}{
          break\tcc*[r]{stop the ray}
        }\uElseIf{$\mathbf{m}=\mathtt{DRILLING}$}{
          $\mathbf{O}[v]\gets \mathtt{FREE}$\tcc*[r]{let ray through}
        }
        \tcc*[h]{Do nothing under the na\"ive mode}
      }
      \eIf{$v = v_{\mathbf{q}}$}{
        $\mathbf{O}[v] \gets \mathtt{OCCUPIED}$ \;
      }{
        $\mathbf{O}[v] \gets \mathtt{FREE}$ \;
      }
    }
  }
  \caption{Raycasting with Augmented Objects}
  \label{alg:raycast}
\end{algorithm}


\begin{figure*}[t]
  \centering
  \includegraphics[width=.225\linewidth]{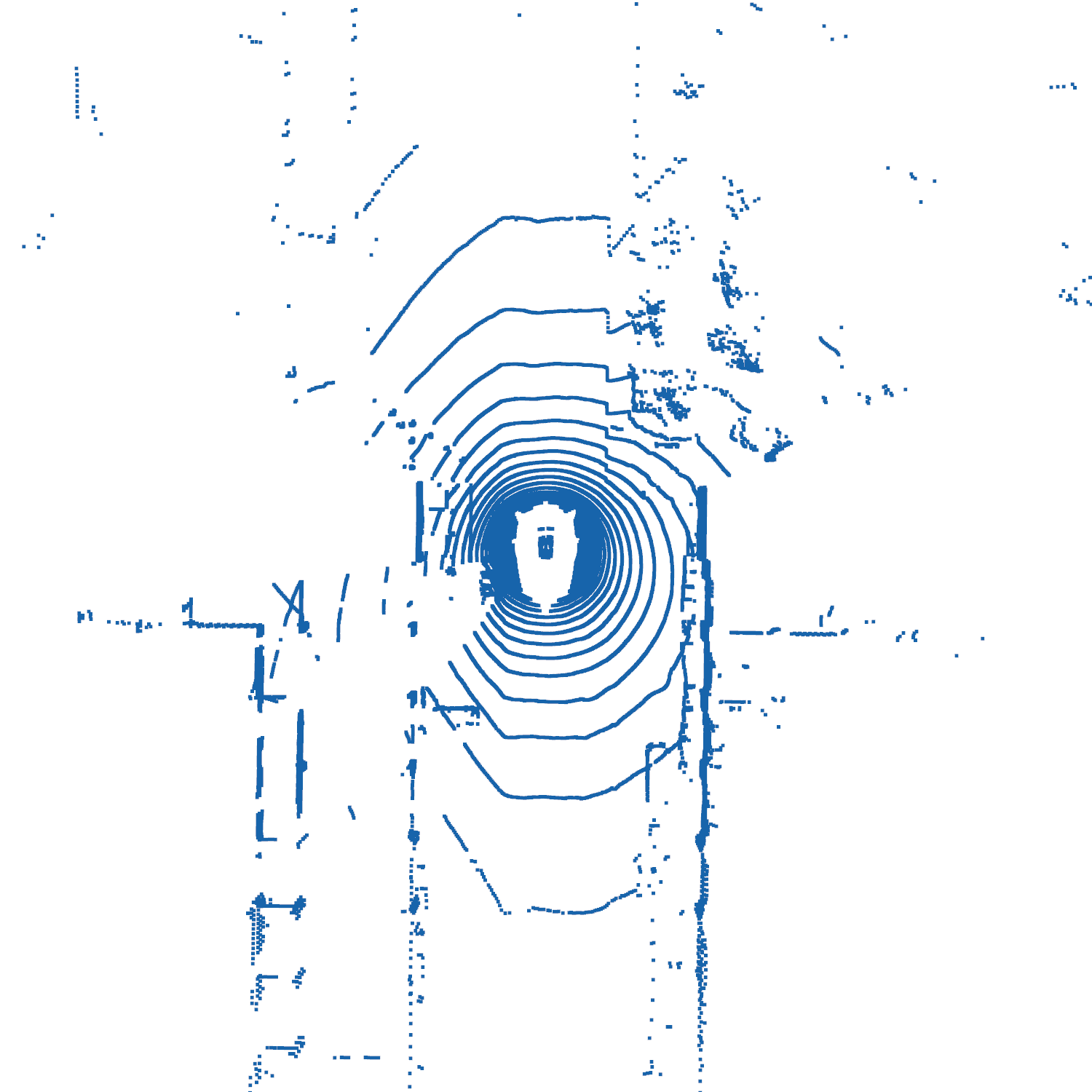}
  \includegraphics[width=.225\linewidth]{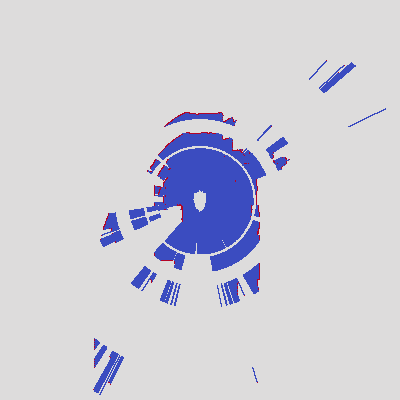}
  \includegraphics[width=.225\linewidth]{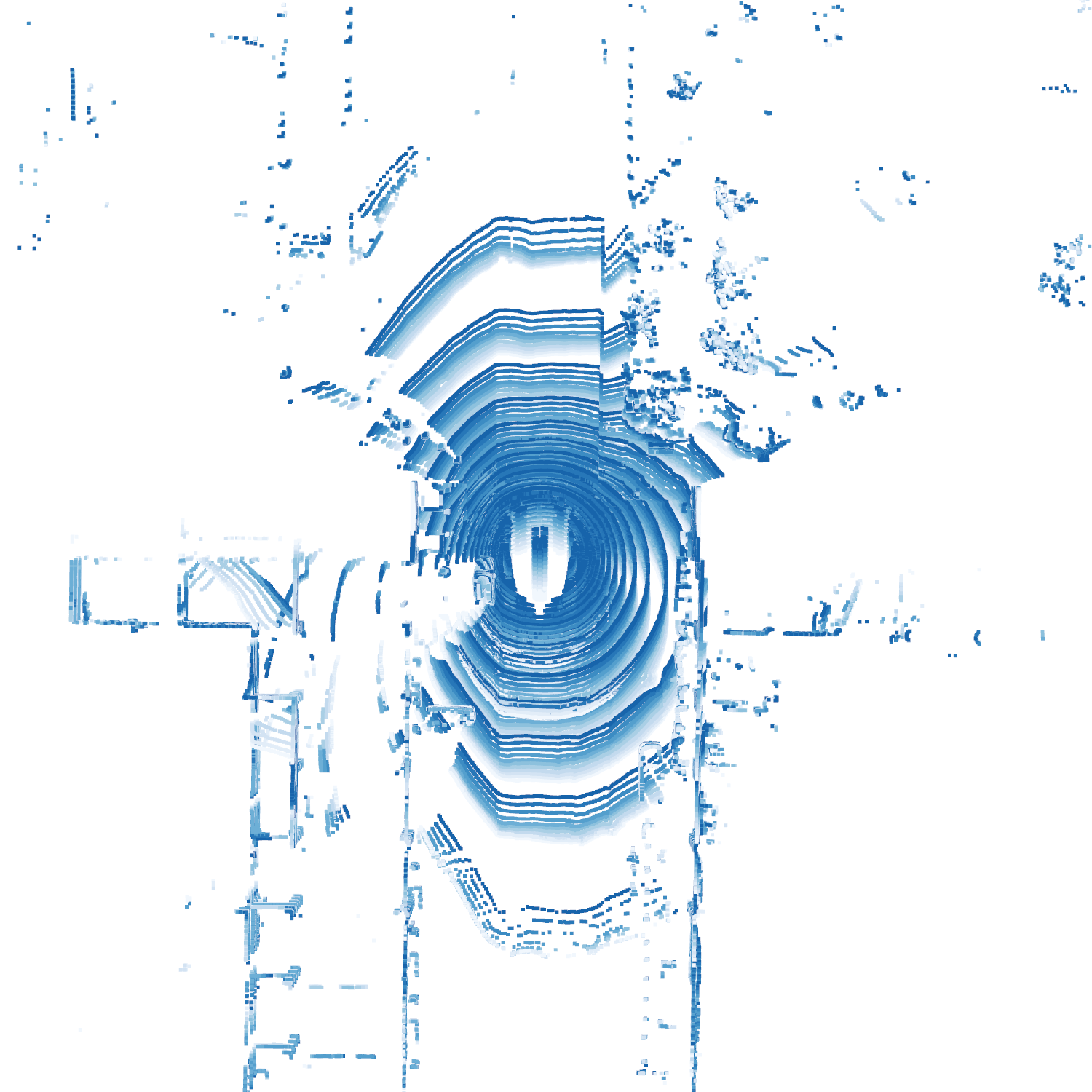}
  \includegraphics[width=.225\linewidth]{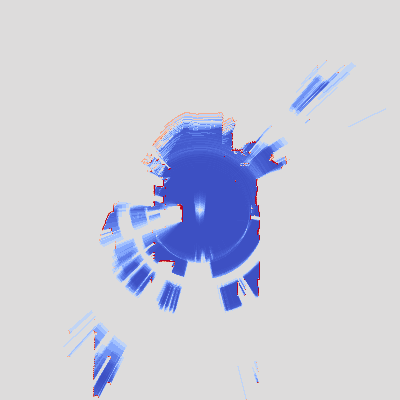}\\
  (a) instantaneous visibility \hspace{13em} (b) temporal occupancy\\
  \caption{We visualize instantaneous visibility vs. temporal occupancy. We choose one xy-slice in the middle to visualize.  Each pixel represents a voxel on the slice. On the \textbf{left}, we visualize a single LiDAR sweep and the instantaneous visibility, which consists of three discrete values: occupied (red), unknown (gray), and free (blue). On the \textbf{right}, we visualize aggregated LiDAR sweeps plus temporal occupancy, computed through Bayesian Filtering~\cite{hornung2013octomap}. Here, the color encodes the probability of the corresponding voxel being occupied: redder means more occupied. }
  \label{fig:slices}
\end{figure*}

{\bf Online occupancy mapping:} How do we extend instantaneous visibility into a temporal context? 
Assume knowing the sensor origin at each timestamp, we can compute instantaneous visibility over every sweep, resulting in 4D spatial-temporal visibility. If we directly integrate a 4D volume into the detection framework, it would be too expensive. 
We seek out online occupancy mapping~\cite{thrun2005probabilistic, hornung2013octomap} and apply Bayesian filtering to turn a 4D spatial-temporal visibility into a 3D posterior probability of occupancy. In Fig.~\ref{fig:slices}, we plot a visual comparison between instantaneous visibility and temporal occupancy. We follow Octomap~\cite{hornung2013octomap}'s formulation and use their off-the-shelf hyper-parameters, e.g. the log-odds of observing freespace and occupied space.


\begin{figure*}
  \centering
  \includegraphics[width=.45\linewidth]{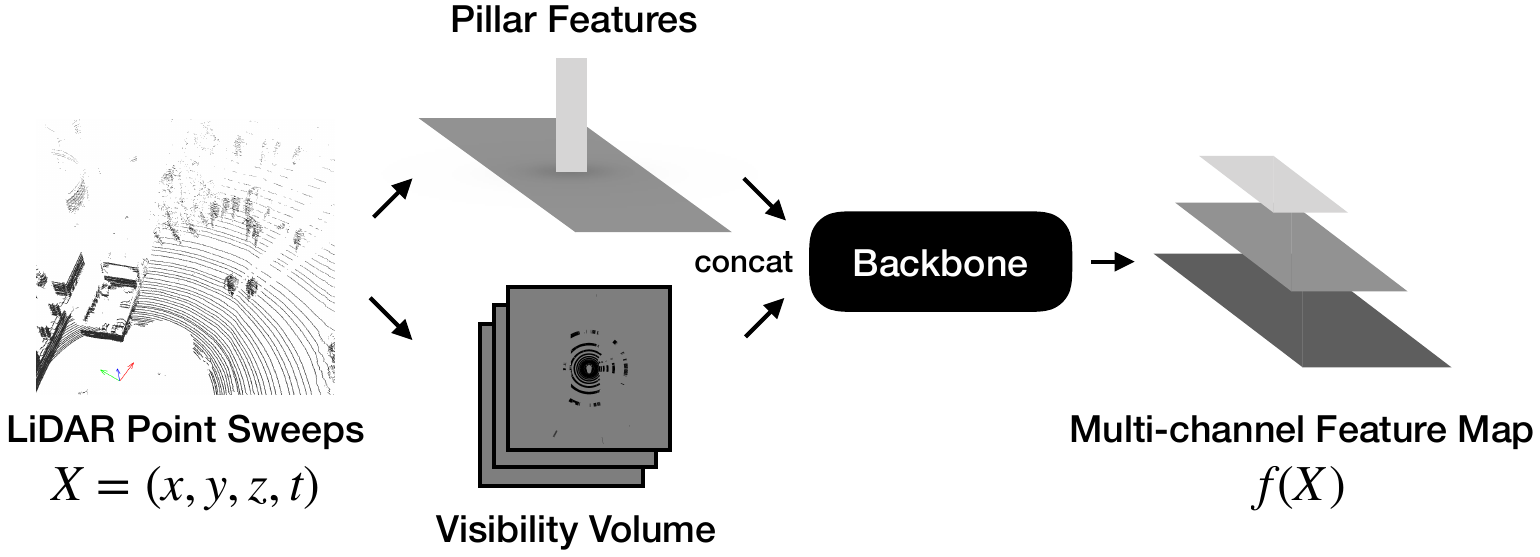}
  \hspace{1em}
  \includegraphics[width=.45\linewidth]{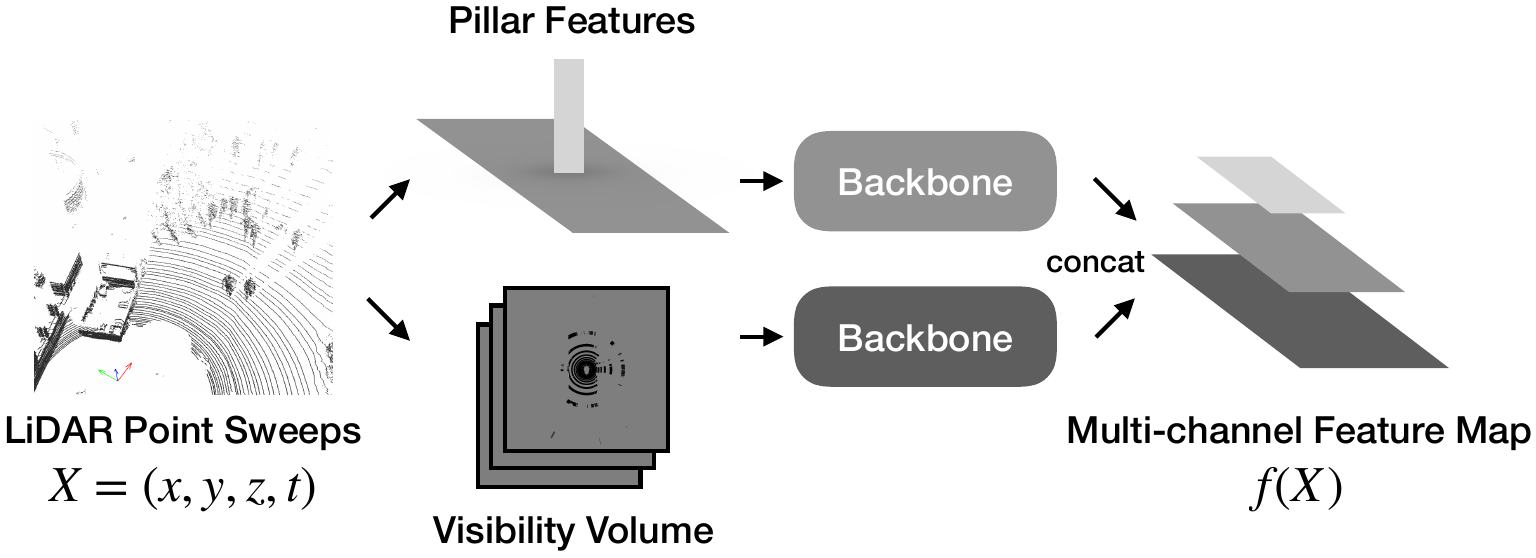}\\
  (a) early fusion \hspace{20em} (b) late fusion
  \caption{We explore both early fusion and late fusion when integrating visibility into the PointPillars model. In the early fusion (a), we concatenate visibility volume with pillar features before applying a backbone network for further encoding. For late fusion, we build one separate backbone network for each stream and concatenate the output of each stream into a final multi-channel feature map. We compare these two alternatives in ablation studies (Tab.~\ref{tab:ablation}). }
  \label{fig:fusion}
\end{figure*}

\subsection{Approach: A Two-stream Network}
Now that we have discussed raycasting approaches for computing visibility, we introduce a novel two-stream network for 3D object detection. We incorporate visibility into a state-of-the-art 3D detector, i.e. PointPillars, as an additional stream. The two-stream approach leverages both the point cloud and the visibility representation and fuses them into a multi-channel representation. We explore both early and late fusion strategies, as illustrated in Fig.~\ref{fig:fusion}. This is a part of the overall architecture illustrated in Fig.~\ref{fig:framework}.

{\bf Implementation:} We implement our two-stream approach by adding an additional input stream to PointPillars. We adopt PointPillars's resolution for discretization in order to improve ease of integration. As a result, our visibility volume has the same 2D spatial size as the pillar feature maps. A simple strategy is to concatenate and feed them into a backbone network. We refer to this strategy as early fusion (Fig.~\ref{fig:fusion}-(a)). Another strategy is to feed each into a separate backbone network, which we refer to as late fusion (Fig.~\ref{fig:fusion}-(b)). We discuss more training details in the Appendix
~\ref{sec:more-details}
. Our code is available online\footnote{\url{https://www.cs.cmu.edu/~peiyunh/wysiwyg}}. 

\section{Experiments}

\begin{table*}[t]
  \centering
  \caption{3D detection mAP on the NuScenes test set. } 
  \label{tab:nuscenes-test}
  \resizebox{.8\linewidth}{!}{
    \begin{tabular}{@{}lcccccccccccc@{}}
      \toprule
      & car & pedes. & barri. & traff. & truck & bus & trail. & const. & motor. & bicyc. & mAP \\ 
      \midrule
      PointPillars~\cite{caesar2019nuscenes} & 68.4 & 59.7 & \textbf{38.9} & \textbf{30.8} & 23.0 & 28.2 & 23.4 & 4.1 & \textbf{27.4} & \textbf{1.1} & 30.5 \\
      Ours & \textbf{79.1} & \textbf{65.0} & 34.7 & 28.8 & \textbf{30.4} & \textbf{46.6} & \textbf{40.1} & \textbf{7.1} & 18.2 & 0.1 & \textbf{35.0} \\
      \bottomrule
    \end{tabular}
  }
\end{table*}

\begin{figure*}
  \centering
  \includegraphics[width=\linewidth]{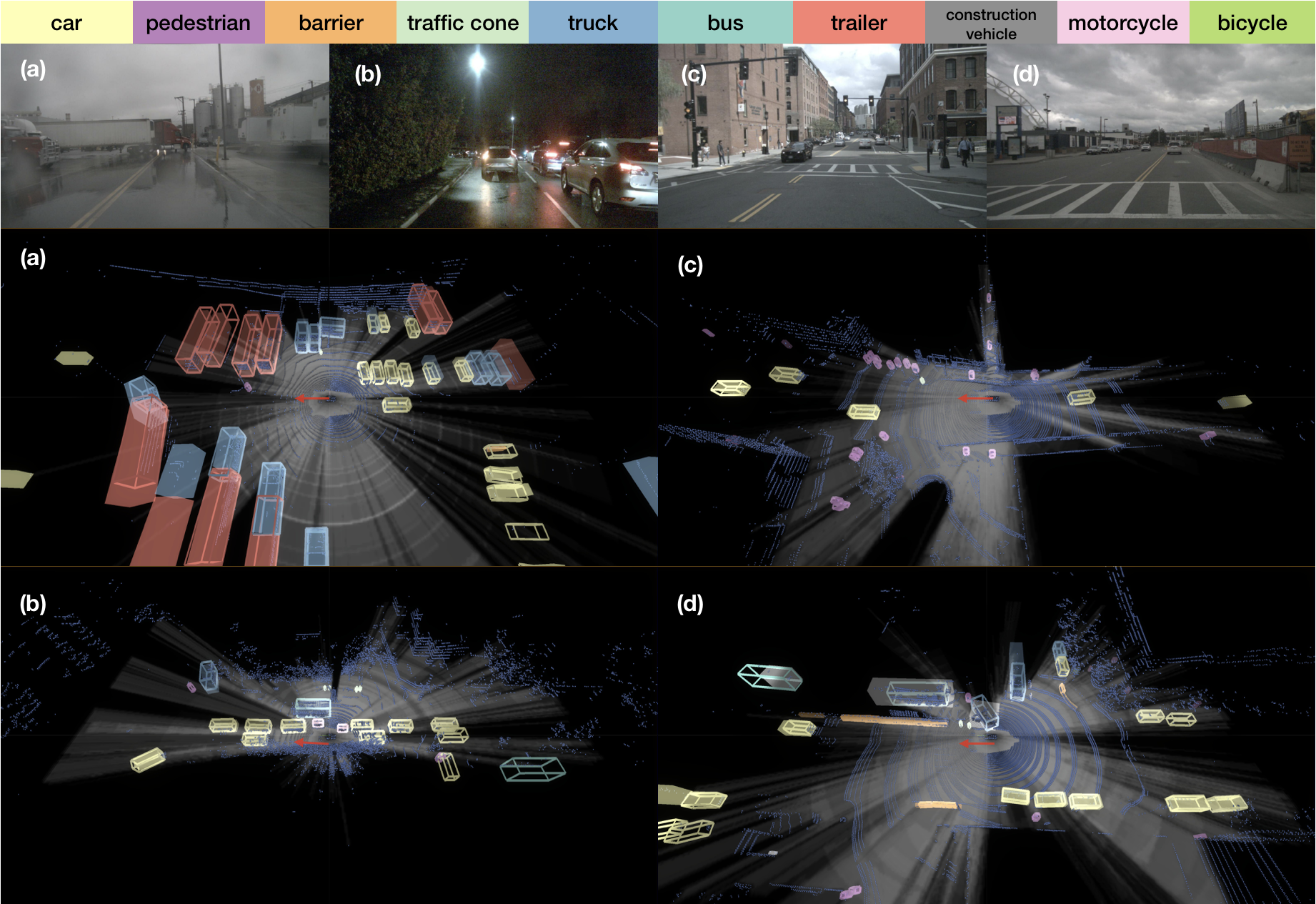}
  \caption{We visualize qualitative results of our two-stream approach on the NuScenes test set. We assign each class a different color (top). We use solid cuboids to represent ground truth objects and wireframe boxes to represent predictions. To provide context, we also include an image captured by the front camera in each scenario. Note the image is \textbf{not} used as part of input for our approach. In (a), our approach successfully detects most vehicles in the scene on a rainy day, including cars, trucks, and trailers. In (b), our model manages to detect all the cars around and also two motorcycles on the right side. In (c), we visualize a scene with many pedestrians on the sidewalk and our model is able to detect most of them. Finally, we demonstrate a failure case in (d), where our model fails to detect objects from rare classes. In this scenario, our model fails to detect two construction vehicles on the car's right side, reporting one as a truck and the other as a bus.} 
  \label{fig:examples}
\end{figure*}

We present both qualitative (Fig.~\ref{fig:examples}) and quantitative results on the NuScenes 3D detection benchmark. We first introduce the setup and baselines, before we present main results on the test benchmark. Afterwards, we perform diagnostic evaluation and ablation studies to pinpoint where improvements come from. Finally, we discuss the efficiency of computing visibility through raycasting on-the-fly.

{\bf Setup:} We benchmark our approach on the NuScenes 3D detection dataset. The dataset contains 1,000 scenes captured in two cities. We follow the official protocol for NuScenes detection benchmark. The training set contains 700 scenes (28,130 annotated frames). The validation set contains 150 scenes (6,019 annotated frames). Each annotated frame comes with one LiDAR point cloud captured by a 32-beam LiDAR, as well as up to 10 frames of (motion-compensated) point cloud. We follow the official evaluation protocol for 3D detection~\cite{caesar2019nuscenes} and evaluate average mAP over different classes and distance threshold.


{\bf Baseline:} PointPillars~\cite{lang2019pointpillars} achieves the best accuracy on the NuScenes detection leaderboard among all published methods that have released source code. The official PointPillars codebase\footnote{\url{https://github.com/nutonomy/second.pytorch}} only contains an implementation for KITTI~\cite{geiger2013vision}. To reproduce PointPillars's results on NuScenes, the authors of PointPillars recommend a third-party implementation\footnote{\url{https://github.com/traveller59/second.pytorch}}.
Using an off-the-shelf configuration provided by the third-part implementation, we train a PointPillars model for 20 epochs from scratch on the full training set and use it as our baseline. This model achieves an overall mAP of 31.5\% on the validation set, which is 2\% higher than the official PointPillars mAP (29.5\%)~\cite{caesar2019nuscenes} (Tab.~\ref{tab:diagnostics-vis}). As suggested by~\cite{caesar2019nuscenes}, the official implementation of PointPillars employ pretraining (ImageNet/KITTI). There is no pretraining in our re-implementation.



{\bf Main results:} We submitted the results of our two-stream approach to the NuScenes test server. In Tab.~\ref{tab:nuscenes-test}, we compare our test-set performance to PointPillars on the official leaderboard~\cite{caesar2019nuscenes}. By augmenting visibility, our proposed approach achieves a significant improvement over PointPillars in overall mAP by a margin of 4.5\%. Specifically, our approach outperforms PointPillars by 10.7\% on cars, 5.3\% on pedestrians, 7.4\% on trucks, 18.4\% on buses, and 16.7\% on trailers. Our model underperforms official PointPillars on motorcycles by a large margin. We hypothesize this might be due to us (1) using a coarser xy-resolution or (2) not pretraining on ImageNet/KITTI. 

\begin{table*}[t]
  \centering
  \caption{3D detection mAP on the NuScenes validation set. \\
  {\footnotesize $^\dagger$: reproduced based on an author-recommended third-party implementation.}
  }
  \label{tab:diagnostics-vis}
  \resizebox{.8\linewidth}{!}{
  \begin{tabular}{@{}lccccccccccc@{}}
    \toprule
    & car & pedes. & barri. & traff. & truck & bus & trail. & const. & motor. & bicyc. & mAP \\
    \midrule
    PointPillars~\cite{caesar2019nuscenes} & 70.5 & 59.9 & 33.2 & \textbf{29.6} & 25.0 & 34.4 & 16.7 & 4.5 & \textbf{20.0} & \textbf{1.6} & 29.5 \\
    PointPillars$^\dagger$ & 76.9 & 62.6 & 29.2 & 20.4 & 32.6 & 49.6 & 27.9 & 3.8 & 11.7 & 0.0 & 31.5 \\
    Ours & \textbf{80.0} & \textbf{66.9} & \textbf{34.5} & 27.9 & \textbf{35.8} & \textbf{54.1} & \textbf{28.5} & \textbf{7.5} & 18.5 & 0.0 &  \textbf{35.4} \\
    \bottomrule
  \end{tabular}
  }
  \\[1em]
  \resizebox{.8\linewidth}{!}{
  \begin{tabular}{@{}lcccc@{}}
    \toprule
    \textit{car} & 0-40\% & 40-60\% & 60-80\% & 80-100\% \\
    \midrule
    Proportion & 20\% & 12\% & 15\% & 54\% \\
    \midrule
    PointPillars$^\dagger$ & 27.2 & 40.0 & 57.2 & 84.3 \\
    Ours & 32.1 & 42.6 & 60.6 & 86.3 \\
    \midrule
    Improvement & \textbf{4.9} & 2.6 & 3.4 & 2.0 \\
    \bottomrule
  \end{tabular}
  ~~
  \begin{tabular}{@{}lcccc@{}}
    \toprule
    \textit{pedestrian} & 0-40\% & 40-60\% & 60-80\% & 80-100\% \\
    \midrule
    Proportion & 20\% & 12\% & 15\% & 54\% \\
    \midrule
    PointPillars$^\dagger$ & 17.3 & 23.4 & 28.0 & 68.3 \\
    Ours & 22.1 & 27.8 & 34.2 & 71.5 \\
    \midrule
    Improvement & 4.8 & 4.4 & \textbf{6.2} & 3.2 \\
    \bottomrule
  \end{tabular}
  }
\end{table*}

{\bf Improvement at different levels of visibility:} We compare our two-stream approach to PointPillars on the validation set, where visibility improves overall mAP by 4\%.  We also evaluate each object class at different levels of visibility. Here, we focus on the two most common classes: car and pedestrian. Interestingly, we observe the biggest improvement over heavily occluded cars (0-40\% visible) and the smallest improvement over fully-visible cars (80-100\% visible). For pedestrian, we also find the smallest improvement is over fully-visible pedestrians (3.2\%), which is 1-3\% less than the improvement over pedestrians with heavier occlusion.

\begin{table*}
\centering
\caption{Ablation studies on the NuScenes validation set. We \textit{italicize} classes for which we perform object augmentation. OA stands for object augmentation and TA stands for temporal aggregation.  }
\label{tab:ablation}
\resizebox{\linewidth}{!}{
\begin{tabular}{l|ccc|ccccccccccc}
    \toprule
    & Fusion & OA & TA & car & pedes. & barri. & traff. & \textit{truck} & \textit{bus} & \textit{trail.} & const. & motor. & bicyc. & avg \\
    \midrule
    (a) & Early & Drilling & Multi-frame & 80.0 & 66.9 & 34.5 & 27.9 & 35.8 & 54.1 & 28.5 & 7.5 & 18.5 & 0.0 & 35.4\\
    (b) & Late & Drilling & Multi-frame & 77.8 & 65.8 & 32.2 & 24.2 & 33.7 & 53.0 & 30.6 & 4.1 & 18.8 & 0.0 & 34.0\\
    (c) & Late & Culling & Multi-frame & 78.3 & 66.4 & 33.2 & 27.3 & 33.4 & 41.6 & 25.7 & 5.6 & 17.0 & 0.1 & 32.9\\
    (d) & Late & Naive & Multi-frame & 78.2 & 66.0 & 32.7 & 25.6 & 33.6 & 51.1 & 27.5 & 4.7 & 15.0 & 0.1 & 33.5\\
    (e) & Late & N/A & Multi-frame & 77.9 & 66.8 & 31.3 & 22.3 & 31.2 & 39.3 & 22.7 & 5.2 & 15.5 & 0.6 & 31.3 \\
    (f) & Late & N/A & Single-frame & 67.9 & 45.7 & 24.0 & 12.4 & 22.6 & 29.9 & 8.5 & 1.3 & 7.1 & 0.0 & 21.9 \\
    (g) & No V & N/A & Single-frame & 68.0 & 38.2 & 20.7 & 8.7 & 23.7 & 28.7 & 11.0 & 0.6 & 5.6 & 0.0 & 20.5 \\
    (h) & Only V & N/A & Single-frame & 66.7 & 28.6 & 15.8 & 4.4 & 17.0 & 25.4 & 6.7 & 0.0 & 1.3 & 0.0 & 16.6 \\
    \midrule
    (i) & No V & Naive & Single-frame & 69.7 & 38.7 & 22.5 & 11.5 & 28.1 & 40.7 & 21.8 & 1.9 & 4.7 & 0.0 & 24.0 \\
    (j) & No V & N/A & Multi-frame & 77.7 & 61.6 & 26.4 & 17.2 & 31.2 & 38.5 & 24.2 & 3.1 & 11.5 & 0.0 & 29.1 \\
    (k) & No V & Naive & Multi-frame & 76.9 & 62.6 & 29.2 & 20.4 & 32.6 & 49.6 & 27.9 & 3.8 & 11.7 & 0.0 & 31.5 \\
    \bottomrule
\end{tabular}
}
\end{table*}

{\bf Ablation studies:} To understand how much improvement each component provides, we perform ablation studies by starting from our final model and removing one component at a time. Key observations from Tab.~\ref{tab:ablation} are:
\setlist{nolistsep}
\begin{itemize}[noitemsep]
\item \textbf{Early fusion} (\textbf{a},b): Replacing early fusion (a) with late fusion (b) results in a 1.4\% drop in overall mAP.
\item \textbf{Drilling} (\textbf{b},c,d): Replacing drilling (b) with culling (c) results in a 11.4\% drop on bus and a 4.9\% drop on trailer. In practice, most augmented trucks and trailers tend to be severely occluded and are removed if the culling strategy is applied. Replacing drilling (b) with naive augmentation (d) results in a 1.9\% drop on bus and 3.1\% drop on trailer, likely due to inconsistent visibility when naively augmenting objects.
\item \textbf{Object augmentation} (\textbf{b},e): Removing object augmentation (b$\rightarrow$e) leads to significant drops in mAP on classes affected by object augmentation, including in a 2.5\% drop on truck, 13.7\% on bus, and 7.9\% on trailer.
\item \textbf{Temporal aggregation} (\textbf{e},f): Removing temporal aggregation (e$\rightarrow$f) leads to worse performance for every class and a 9.4\% drop in overall mAP.
\item \textbf{Visibility stream} (\textbf{f},g,h): Removing the visibility stream off a \textit{vanilla} two-stream approach (f$\rightarrow$g) drops overall mAP by 1.4\%. Interestingly, the most dramatic drops are over pedestrian (+7.5\%), barrier(+3.3\%), and traffic cone (+3.7\%). Shape-wise, these objects are all ``skinny'' and tend to have less LiDAR points on them. This suggests visibility helps especially when having less points. The network with only a visibility stream (h) underperforms a \textit{vanilla} PointPillars (g) by 4\%. 
\item \textbf{Vanilla PointPillars} (\textbf{g},i,j,k): On top of vanilla PointPillars, object augmentation (g$\rightarrow$i) improves mAP over augmented classes by 9.1\%; temporal aggregation (g$\rightarrow$j) improves overall mAP by 8.6\%. Adding both (g$\rightarrow$k) improves overall mAP by 11.0\%.
\end{itemize}

{\bf Run-time speed:} We implement visibility computation in C++ and 
integrate it into PyTorch training as part of (parallel) data loading. On an Intel i9-9980XE CPU, it takes 24.4$\pm$3.5ms on average to compute visibility for a 32-beam LiDAR point cloud when running on a single CPU thread.

{\bf Conclusions:} We revisit the problem of finding a good representation for 3D data. We point out that contemporary representations are designed for true 3D data (e.g. sampled from mesh models). In fact, 3D sensored data such as a LiDAR sweep is 2.5D. By processing such data as a collection of \textit{normalized} points $(x,y,z)$, important visibility information is fundementally destroyed. In this paper, we augment visibility into 3D object detection. We first demonstrate that visibility can be efficiently re-created through 3D raycasting. We introduce a simple two-stream approach that adds visibility as a separate stream to an existing state-of-the-art 3D detector. We also discuss the role of visibility in placing virtual objects for data augmentation and explore visibility in a temporal context - building a local occupancy map in an online fashion. Finally, on the NuScenes detection benchmark, we demonstrate that the proposed network outperforms state-of-the-art detectors by a significant margin.

{\bf Acknowledgments:} This work was supported by the CMU Argo AI Center for Autonomous Vehicle Research.

\appendix
\setcounter{section}{0}
\setcounter{figure}{0}
\setcounter{footnote}{0}
\renewcommand{\thesection}{A\arabic{section}}
\renewcommand{\thefigure}{\Alph{figure}}
\def\thesection{\Alph{section}}

\section{Additional Results}
\label{sec:additional-results}

\begin{figure*}[ht]
  \centering
  \includegraphics[width=\linewidth]{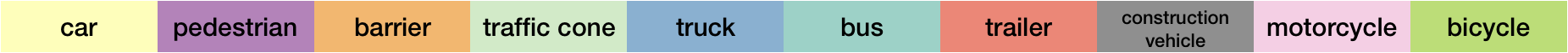}
  \includegraphics[width=\linewidth]{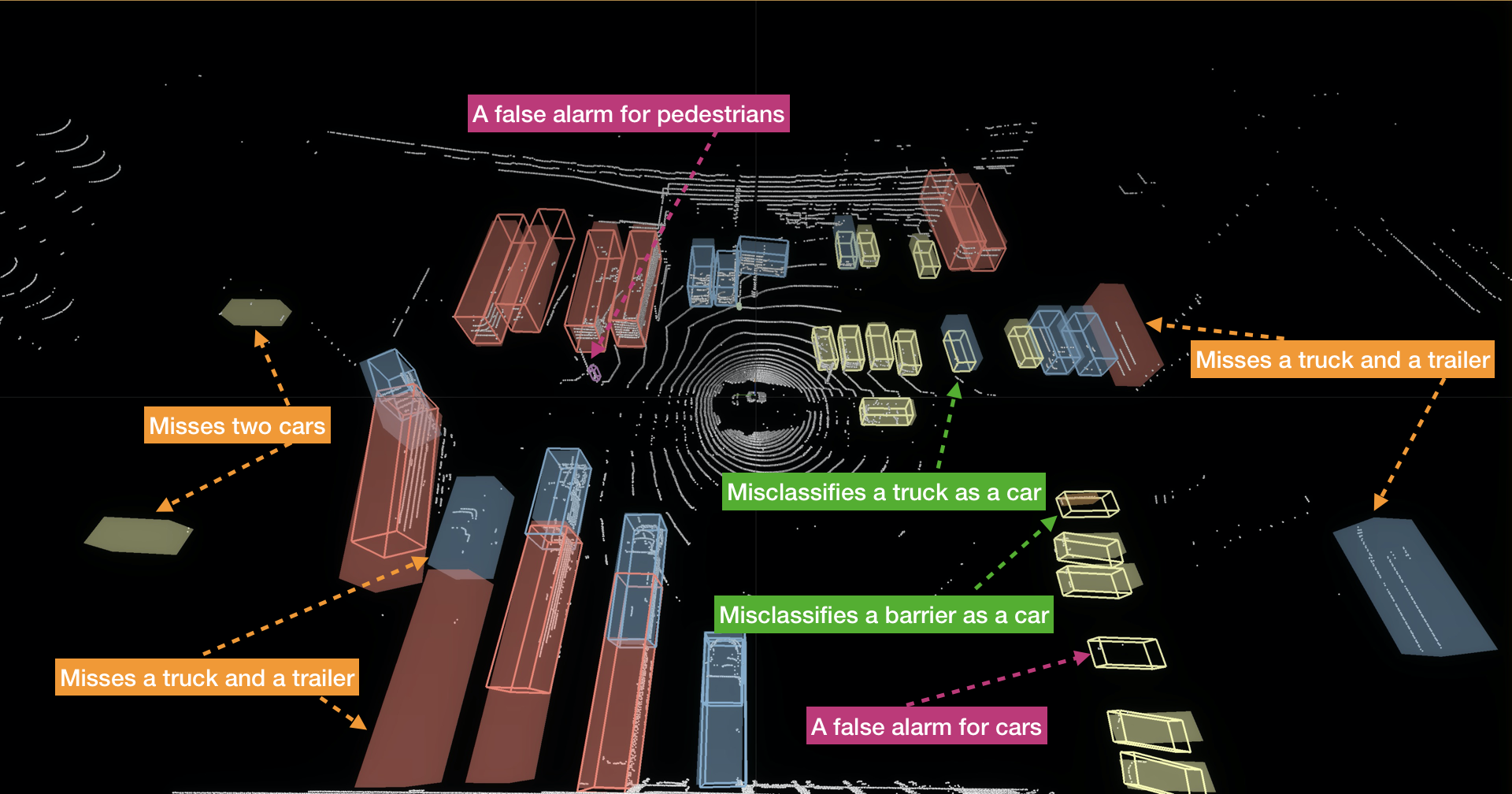}
  \caption{We use an example frame (same as Fig.
    ~\ref{fig:examples}-(a)) from the video to illustrate how we visualize. Solid transparent cuboids represent ground truth. Wireframe boxes represent predictions. Colors encode object classes (top). Inside this frame, our algorithm successfully detects most cars, trucks, and trailers. We also highlight all the mistakes made by our algorithm. }
  \label{fig:video}
\end{figure*}

{\bf More qualitative examples:} Please find a result video at \url{https://youtu.be/8bXkDxSgMsM}. We provide a visual guide on how to interpret the visualization in Fig~\ref{fig:video}.


{\bf Performance over multiple runs:} We re-train the same model for 5 more times, each time with a different random seed. The seed affects both network initialization and the order in which training data is sampled. We evaluate the detection accuracy of each run on nuScenes validation set and report the mean and standard deviation. As Table~\ref{tab:val} shows, the standard deviations are much smaller than the reported improvements, suggesting the performance gain is not due to a lucky run.
\begin{table*}
  \centering
  \caption{3D detection mAP over multiple runs on NuScenes validation set. \\
    {\footnotesize $^\dagger$: reproduced based on an author-recommended third-party implementation.}
  }
  \label{tab:val}
  \resizebox{\textwidth}{!}{
    \begin{tabular}{@{}cccccccccccc@{}}
      \toprule
      & car & pedes. & barri. & traff. & truck & bus & trail. & const. & motor. & bicyc. & mAP \\
      \midrule
      PointPillars$^\dagger$ & 76.9 & 62.6 & 29.2 & 20.4 & 32.6 & 49.6 & 27.9 & 3.8 & 11.7 & 0.0 & 31.5 \\
      Ours & 80.0 & 66.9 & 34.5 & \textbf{27.9} & 35.8 & 54.1 & 28.5 & 7.5 & \textbf{18.5} & 0.0 & 35.4 \\
      \midrule
      Ours (run1) & 80.2 & 67.3 & 34.5 & 27.0 & 36.1 & \textbf{54.6} & 28.9 & 6.4 & 17.5 & \textbf{0.3} & 35.3 \\
      Ours (run2) & 80.5 & 67.1 & 35.2 & 27.0 & 36.1 & 52.9 & 30.5 & 6.9 & 16.2 & 0.0 & 35.2 \\
      Ours (run3) & 80.7 & 66.9 & 35.4 & 24.9 & 35.0 & 54.0 & \textbf{31.7} & 6.8 & 16.6 & 0.1 & 35.2 \\
      Ours (run4) & \textbf{80.8} & 67.2 & 34.5 & \textbf{27.9} & 35.8 & 52.9 & 29.6 & \textbf{8.2} & 14.8 & 0.0 & 35.2 \\
      Ours (run5) & 80.4 & \textbf{67.7} & \textbf{35.6} & 26.8 & \textbf{36.7} & 52.3 & 31.1 & 7.9 & 17.8 & 0.0 & \textbf{35.6} \\
      Mean $\pm$ Std & 80.4 $\pm$ 0.3 & 67.2 $\pm$ 0.3 & 35.0 $\pm$ 0.5 & 26.9 $\pm$ 1.1 & 35.9 $\pm$ 0.6 & 53.5 $\pm$ 0.9 & 30.1 $\pm$ 1.3 & 7.3 $\pm$ 0.7 & 16.9 $\pm$ 1.3 & 0.1 $\pm$ 0.1 & 35.3 $\pm$ 0.2 \\
      \bottomrule
    \end{tabular}
  }

\end{table*}

\section{Additional Experimental Details}
\label{sec:more-details}
Here, we provide additional details about our method, including pre-processing, network structure, initialization, loss function, training etc. These details apply to both the baseline method (PointPillars) and our two-stream approach.

{\bf Pre-processing:} We focus on points whose $(x,y,z)$ satisfies $x\in [-50,50], y\in [-50,50], z\in [-5, 3]$ and ignore points outside the range when computing pillar features. We group points into vertical columns of size $0.25\times0.25\times8$. We call each vertical column a pillar. We resample to make sure each non-empty pillar contains 60 points. For raycasting, we do \textbf{not} ignore points outside the range and use a voxel size of $0.25\times0.25\times0.25$.

{\bf Network structure:} We introduce (1) pillar feature network; (2) backbone network; (3) detection heads.
\begin{enumerate}[label={(\arabic*)}]
\item Pillar feature network operates over each non-empty pillar. It takes points $(x,y,z,t)$ within the pillar and produces a 64-d feature vector. To do so, it first compresses $(x,y)$ to $(r)$, where $r=\sqrt{x^2+y^2}$. Then it augments each point with its offset to the pillar's arithmetic mean $(x_c,y_c,z_c)$ and geometric mean $(x_p,y_p)$. Please refer to Sec. 2.1 of PointPillars~\cite{lang2019pointpillars} for more details. Then, it processes augmented points $(r,z,t,x-x_c,y-y_c,z-z_c,x-x_p,y-y_p)$ with a 64-d linear layer, followed by BatchNorm, ReLU, and MaxPool, which results in a 64-d embedding for each non-empty pillar. Conceptually, this is equivalent to a mini one-layer PointNet. Finally, we fill empty pillars with all zeros. Based on how we discretize, pillar feature network produces a $400\times400\times64$ feature map.
\item Backbone network is a convolutional network with an encoder-decoder structure. This is also sometimes referred to as Region Proposal Network. Please read VoxelNet~\cite{zhou2018voxelnet}, SECOND~\cite{yan2018second}, and PointPillars~\cite{lang2019pointpillars} for more details. The network consists three blocks of fully convolutional layers. Each block consists of a convolutional stage and a deconvolutional stage. The first (de)convolution filter of the (de)convolutional stage changes the spatial resolution and the feature dimension. All (de)convolution is 3x3 and followed with BatchNorm and ReLU. For our two-stream early-fusion model, the backbone network takes an input of size $400\times400\times96$, where $64$ channels are from pillar feature and $32$ channels are from visibility. The first block contains 4 convolutional layers and 4 deconvolutional layers. The second and the third block each consists of 6 both of these layers. Within the first block, the feature dimension changes from $400\times400\times96$ to $200\times200\times96$ during the convolutional stage, and $200\times200\times96$ to $100\times100\times192$ during the deconvolutional stage. Within the second block, the feature dimension from $200\times200\times96$ to $100\times100\times192$. Within the third block, the feature map changes from $100\times100\times192$ to $50\times50\times384$ and back to $100\times100\times192$. At last, features from all three blocks are concatenated as the final output, which has a size of $100\times100\times576$.
\item Detection heads include one for large object classes (i.e. car, truck, trailer, bus, and construction vehicles) and one for small object classes (i.e. pedestrian, barrier, traffic cone, motorcycle, and bicycle). The large head takes the concatenated feature map from backbone network as input ($100\times100\times576$) while the small head takes the feature from the backbone's first convolutional stage as input ($200\times200\times96$). Each head contains a linear predictor for anchor box classification and a linear prediction for bounding box regression. The classification predictor outputs a confidence score for each anchor box and the regression predictor outputs adjustment coefficients (i.e. $x, y, z, w, l, h, \theta$).
\end{enumerate}

{\bf Loss function:} For classification, we adopt focal loss~\cite{lin2017focal} and set $\alpha=0.25$ and $\gamma=2.0$. For regression output, we use smooth L1 loss (a.k.a. Huber loss) and set $\sigma=3.0$, where $\sigma$ controls where the transition between L1 and L2 happens. The final loss function is the classification loss multiplied by 2 plus the regression loss.

{\bf Training:} We train all of our models for 20 epochs and optimize using Adam~\cite{kingma2014adam} as the optimizer. We follow a learning rate schedule known as ``one-cycle''~\cite{smith2017cyclical}. The schedule consists of 2 phases. The first phase includes the first 40\% training steps, during which we increase the learning rate from $\frac{0.003}{10}$ to 0.003 while decreasing the momentum from 0.95 to 0.85 following cosine annealing. The second phase includes the rest 60\% training steps, during which we decrease the learning rate from 0.003 to $\frac{0.003}{10000}$ while increasing the momentum from 0.85 to 0.95. We use a fixed weight decay of 0.01.

{\small
  \bibliographystyle{ieee_fullname}
  \bibliography{ref}
}

\end{document}